\documentclass[11pt,a4paper]{article}
\usepackage{emnlp2021}
\usepackage{times}
\usepackage{latexsym}

\usepackage[T5,T2A,T1]{fontenc}
\usepackage[utf8]{inputenc}

\usepackage{microtype}

\usepackage{CJKutf8}
\usepackage{tabularx}
\usepackage{booktabs}
\usepackage{graphicx}
\usepackage{subfigure}
\usepackage{caption}
\usepackage{xspace}
\usepackage{colortbl}
\usepackage{xcolor}
\usepackage{framed}
\usepackage{footnote}
\makesavenoteenv{tabular}
\makesavenoteenv{table}
\usepackage{amsfonts}
\usepackage{multirow}
\definecolor{Gray}{gray}{0.9}

\newcommand{\resource}{\textsc{AM$^2$iCo}\xspace}
\newcommand{\ignore}[1]{}

\title{\resource: Evaluating Word Meaning in Context across Low-Resource Languages with Adversarial Examples}

\author{Qianchu Liu$^1$,~Edoardo M. Ponti$^{1}$ $^2$,~Diana McCarthy$^1$,   ~Ivan Vuli\'{c}$^{1}$,  ~Anna Korhonen$^1$ \smallskip \\
$^1$ Language Technology Lab, TAL, University of Cambridge, UK \\
$^2$ Mila/McGill University,
Montreal, Canada \\
\texttt {\{ql261,ep490,iv250,alk23\}@cam.ac.uk} \\
\texttt {diana@dianamccarthy.co.uk }
}

\begin{document}
\maketitle
\begin{abstract}
Capturing word meaning in context and distinguishing between correspondences and variations across languages is key to building successful multilingual and cross-lingual text representation models. However, existing multilingual evaluation datasets that evaluate lexical semantics ``in-context'' have various limitations. In particular,  \textbf{1)} their language coverage is restricted to high-resource languages and skewed in favor of only a few language families and areas, \textbf{2)} a design that makes the task solvable via superficial cues, which results in artificially inflated (and sometimes super-human) performances of pretrained encoders, 
and \textbf{3)} little support for cross-lingual evaluation. In order to address these gaps, we present \resource (\textbf{A}dversarial and \textbf{M}ultilingual \textbf{M}eaning \textbf{i}n \textbf{Co}ntext), a wide-coverage cross-lingual and multilingual evaluation set; it aims to faithfully assess the ability of state-of-the-art (SotA) representation models to understand the identity of word meaning in cross-lingual contexts for 14 language pairs. We conduct a series of experiments in a wide range of setups and demonstrate the challenging nature of \resource. The results reveal that current SotA pretrained encoders substantially lag behind human performance, and the largest gaps are observed for low-resource languages and languages dissimilar to English. 

\end{abstract}

\section{Introduction}
\label{s:introduction}
Pretrained language models (LMs) such as BERT \cite{devlin2018bert} and XLM-R \cite{conneau-etal-2020-unsupervised} offer a natural way to distinguish different word meanings in context without performing explicit sense disambiguation. This property of ``meaning contextualization'' is typically evaluated either via standard entity linking 
\cite{rao2013entity, shen2014entity} and Word Sense Disambiguation (WSD) tasks \cite{navigli2009word,Moro:2014tacl,Raganato:2017eacl} or, recently, via the Word-in-Context (WiC) evaluation paradigm \cite{pilehvar-camacho-collados-2019-wic,raganato-etal-2020-xl}. 

Although monolingual evaluation in English is still predominant, a need has been recognized to construct similar resources for other languages to support cross-lingual evaluation and model diagnostics. This includes multilingual and cross-lingual WSD benchmarks \cite[\textit{inter alia}]{Navigli:2012emnlp,Navigli:2013semeval,Scarlini:2020lrec,Barba:2020ijcai}, cross-lingual entity linking \cite{tsai-roth-2016-cross,Raiman:2018aaai,Upadhyay:2018emnlp} and, most recently, multilingual WiC (termed XL-WiC) spanning 12 languages \cite{raganato-etal-2020-xl}.

This most recent WiC evaluation approach is particularly attractive as \textbf{1)} it bypasses the dependence on modeling predefined ontologies (entity linking) and explicit sense inventories (WSD), and \textbf{2)} it is framed as a simple binary classification task: for a target word $w$ appearing in two different contexts $c_1$ and $c_2$, the system must decide whether $w$ conveys the same meaning in both contexts, or not. 

However, the current WiC evaluation still allows ample room for improvement: \textbf{1)} current language coverage is limited, and biased towards resource-rich Indo-European languages; 
\textbf{2)} coverage of lexical concepts, due to their paucity in language-specific WordNets, is also limited; 
\textbf{3)} XL-WiC is a monolingual resource available in different languages, i.e., it does not support cross-lingual assessments. Further, \textbf{4)} the current WiC datasets offer low human upper bounds and inflated (even super-human) performance for some languages.\footnote{In turn, this might give a false impression that some languages in the task are 'solved' by the current pretrained LMs.} This is due to superficial cues where \textbf{5)} many examples in the current WiC datasets can be resolved relying either on the target word alone without any context or on the context alone, which eludes evaluation honing in on the interplay between target words and their corresponding contexts.  




In order to address these limitations and provide a more comprehensive evaluation framework, we present \textbf{\resource} (\textbf{A}dversarial and \textbf{M}ultilingual \textbf{M}eaning \textbf{i}n \textbf{Co}ntext), a novel multilingual and cross-lingual WiC task and resource. It covers a typologically diverse set of 15 languages, see Table~\ref{table:data splits}. Based on Wikipedia in lieu of WordNet, \resource covers a wider set of ambiguous words which especially complements WiC on the long tail of entity names and adds challenges in generalization for a larger vocabulary than a restricted set of common words. More  importantly, the use of Wikipedia  enables WiC evaluation on low-resource languages (e.g., Basque, Georgian, Bengali, Kazakh). We also improve the WiC resource design; it now \textbf{1)} includes adversarial examples and careful data extraction procedures to prevent the models from backing off to superficial clues, \textbf{2)} results in a more challenging benchmark with truer and much wider gaps between current SotA pretrained encoders and human capability (see \S\ref{sec:add challenge}), and \textbf{3)} enables cross-lingual evaluation and analysis.

The ample and diverse data in \resource enables a wide spectrum of experiments and analyses in different scenarios. We evaluate SotA pretrained encoders, multilingual BERT and XLM-R, both off-the-shelf using a metric-based approach (i.e., without any task adaptation) and after task-specific fine-tuning. With fine-tuned models, we investigate zero-shot cross-lingual transfer as well as transfer from multiple source languages. In general, our results across these diverse scenarios firmly indicate a large gap between human and system performance across the board, which is even more prominent when dealing with resource-poor languages and languages dissimilar to English, holding promise to guide modeling improvements in the future.

In the hope that \resource will be a challenging and valuable diagnostic and evaluation asset for future work in multilingual and cross-lingual representation learning, we release the data along with the full guidelines at 
\url{https://github.com/cambridgeltl/AM2iCo}.





\section{\resource: Cross-Lingual Word-in-Context Evaluation}

\noindent \textbf{Task Definition.} 
\resource is a standard binary classification task on pairs of word in context instances. Each pair consists of a target word with its context in English and a target word with its context in a target language. Formally, each dataset of \resource spans a set of $N$ examples $\hat{x}_i$, $i=1,\ldots,N$ for a language pair. Each example $\hat{x}_i$ is in fact a pair of items $\hat{x}_i = (x_{i,src}, x_{i,trg})$, where the item $x_{i,src}$ is provided in the source language $L_{src}$ and the item $x_{i,trg}$ is in the target language $L_{trg}$. The item $x_{i,src}$ in turn is another pair $x_{i,src} = (w_{i,src}, c_{i,src})$; it contains a target word $w_{i,src}$ from $L_{src}$ and its (wider) context $c_{i,src}$ (also in $L_{src}$) in which that word appears, see Table~\ref{examp}; the same is valid for $x_{i,trg}$. The classification task is then to judge whether the words $w_{i,src}$ and $w_{i,trg}$ occurring in the respective contexts $c_{i,src}$ and $c_{i,trg}$ have the same sense/meaning (i.e., whether they refer to the same entity/concept), or not.

\vspace{1.4mm}
\noindent \textbf{Final Resource.} The full \resource resource comprises datasets for 14 language pairs, where English is paired with 14 target languages. For brevity, in the rest of the paper we refer to the dataset of each language pair simply with the $L_{trg}$ language code (e.g., ZH instead of EN-ZH); languages and codes are provided in Table~\ref{table:data splits}.

As illustrative examples, we show a positive pair (label `T') and a negative pair (label `F') from the ZH \resource dataset in Table~\ref{examp} (Examples 1 and 2).  In the positive example, both target words `Apollo' and \begin{CJK*}{UTF8}{gbsn}`
阿波罗'\end{CJK*} in their contexts refer to the same concept: the Apollo spaceflight program. In the negative example, the Chinese target word \begin{CJK*}{UTF8}{gbsn}`
阿波罗'\end{CJK*} refers to the Apollo aircraft, but the English target word `Apollo' now refers to the Greek God. 


\begin{table*}[t!]
\centering
{\footnotesize
\def\arraystretch{1.10}
\begin{tabularx}{\textwidth}{l Xp{0.45\textwidth} l}
\toprule
 {\bf no.} & {\bf English $x_{i,src}$}  & {\bf Chinese $x_{i,trg}$} & {\bf Label} \\
 \cmidrule(lr){2-4}
 1&Bill Kaysing ( July 31 , 1922 – April 21 , 2005 ) was an American writer who claimed that the six \colorbox{Gray}{\bf Apollo} Moon landings between July 1969 and December 1972 were hoaxes , and so a founder of the Moon hoax movement .&\begin{CJK*}{UTF8}{gbsn}泰坦系列导弹的发射任务结束后，LC-16被移交给NASA用做双子座计划的航天员训练及 ~\colorbox{Gray}{\bf 阿波罗 }~ 中飞船服务舱的静态试车。[...]
 \end{CJK*}\newline\textit{(After the launch of the Titan missiles, LC-16 was handed over to NASA for the training of the astronauts in the Gemini program and the static test run of the service module of the spacecraft in \colorbox{Gray}{\bf Apollo} [...])} &T\\
  \cmidrule(lr){2-4}
 2&Nearer the house , screening the service wing from view , is a Roman triumphal arch , the `` Temple of \colorbox{Gray}{\bf Apollo} '' , also known ( because of its former use a venue for cock fighting ) as `` Cockpit Arch '' , which holds a copy of the famed Apollo Belvedere . &  \begin{CJK*}{UTF8}{gbsn}阿波罗-联盟测试计划中，美国的 ~\colorbox{Gray}{阿波罗} ~ 航天器和苏联的联盟航天器在地球轨道中对接。...\end{CJK*} \newline \textit{(In the Apollo-Soyuz test plan, America's \colorbox{Gray}{\bf Apollo} spacecraft and the Soviet Union's Soyuz spacecraft are docked in the Earth orbit [...])} &F\\
 \cmidrule(lr){2-4}
 3&Bill Kaysing ( July 31 , 1922 – April 21 , 2005 ) was an American writer who claimed that the six \colorbox{Gray}{\bf Apollo} Moon landings between July 1969 and December 1972 were hoax es , and so a founder of the Moon hoax movement .&\begin{CJK*}{UTF8}{gbsn}泰坦系列导弹的发射任务结束后，LC-16被移交给 ~\colorbox{Gray}{\bf NASA}~ 用做双子座计划的航天员训练及阿波罗中飞船服务舱的静态试车。[...]\end{CJK*}
 \newline \textit{(After the launch of the Titan missiles, LC-16 was handed over to \colorbox{Gray}{\bf NASA} for the training of the astronauts in the Gemini program and the static test run of the service module of the spacecraft in Apollo [...])}
 &F\\
 
 \bottomrule
\end{tabularx}
}%
\caption{Positive (1), negative (2) and adversarial negative examples (3) from ZH \resource. Target words are provided in boldface and with a gray background. Translations of the ZH items are provided in italic.}
\label{examp}
\end{table*}

\begin{table*}
\centering
\def\arraystretch{0.83}
{\small
\begin{tabularx}{\textwidth}{l XXXXXXXXXXXXXX}
\toprule
&DE&	RU&	JA&	ZH&	AR	&KO	&FI&	TR&	ID&	EU&	KA&	BN	&KK&	UR\\
\cmidrule(lr){2-15}
train &50,000&	28,286&	16,142&	13,154&	9,622&	7,070&	6,322&	3,904&	1,598&	978&--&--&--&--\\
dev & 500 &	500 &	500	&500&	500	&500&	500	&500&	500&	500&	500&	332&	276&	108\\	
test &1,000 &	1,000 &	1,000 &	1,000 &	1,000 &	1,000 &	1,000	 &1,000 &	1,000 &	1,000 &	1,000 &	700	 &400 &	400\\

\bottomrule

\end{tabularx}
}%
\caption{Data sizes for \resource across 14 language pairs. We also provide larger dev and test sets for DE and RU spanning 5,000 and 10,000 examples, respectively. \textbf{EN}=English; \textbf{DE}=German; \textbf{RU}=Russian; \textbf{JA}=Japanese; \textbf{ZH}=Chinese; \textbf{AR}=Arabic; \textbf{KO}=Korean; \textbf{FI}=Finnish; \textbf{TR}=Turkish; \textbf{ID}=Indonesian; \textbf{EU}=Basque; \textbf{KA}=Georgian; \textbf{BN}=Bengali, \textbf{KK}=Kazakh; \textbf{UR}=Urdu.}
 \label{table:data splits}
\end{table*}


In what follows we describe the creation of \resource. We also demonstrate the benefits of \resource and its challenging nature. 

\subsection{Data Creation}
\label{s:datacreation}


Wikipedia is a rich source of disambiguated contexts for multiple languages. The availability of Wikipedia's cross-lingual links provides a direct way to identify cross-lingual concept correspondence. The items $x_{i,src}$ and $x_{i,trg}$ are then extracted by taking the surrounding (sentential) context of a hyperlinked word in a Wikipedia article. We balance the context length by (i) discarding items longer than 100 words, and (ii) adding preceding and following sentences to the context for sentences shorter than 30 words. Using the Wikipedia dumps of our 15 languages (see Table~\ref{table:data splits}), we create monolingual items $x$ for each language. We select only \textit{ambiguous} target words ($w$-s), that is, words that link to at least two different Wikipedia pages.\footnote{To avoid rare words that are potentially unknown to non-experts, we retain only words that are among the top 200k words by frequency in each respective Wikipedia.}



For each word, we then create monolingual positive examples by pairing two items (i.e., word-context pairs) $x_i$ and $x_j$ in which the same target word $w$ is linked to the same Wikipedia page, signaling the same meaning. In a similar fashion, monolingual negative examples are created by pairing two items where the same target word $w$ is linked to two different Wikipedia pages. We ensure that there is roughly an equal number of positive and negative examples for each target word. 

Now, each monolingual example (i.e., pair of items) $\hat{x}$ contains the same word occurring in two different contexts. In order to create a cross-lingual dataset, we leverage the Wikipedia cross-lingual links; we simply (i) replace one of the two items from each English pair with an item in the target language, and (ii) replace one of the two items from each target language pair with an English item, where the cross-lingual replacements point to the same Wikipedia page as indicated by the cross-lingual Wiki links. Through this procedure, the final datasets cover a sufficient (and roughly comparable) number of examples containing ambiguous words both in English and in $L_{trg}$. We also rely on data selection heuristics that improve the final data quality, discussed in \S\ref{ss:quality} and \S\ref{sec:add challenge}.


Finally, in each cross-lingual dataset we reserve 1,000 examples for testing, 500 examples as dev data; the rest is used for training. The exception are 4 resource-poor languages, where all the data examples are divided between dev and test. All data portions in all datasets are balanced. We ensure zero overlap between train, dev, and test portions. The final \resource statistics are given in Table~\ref{table:data splits}.


\vspace{1.4mm}
\noindent \textbf{Human Validation.}
We employ human annotators to assess the quality of \resource. For each dataset, we recruit two annotators who each validate a random sample of 100 examples, where 50 examples are shared between the two samples and are used to compute inter-rater agreement.\footnote{The annotators were recruited via two crowdsourcing platforms, Prolific and Proz, depending on target language coverage. The annotators were native speakers of the target language, fluent in English, and with an undergraduate degree.}


\subsection{Data Selection Heuristics}
\label{ss:quality}
One critical requirement of \resource is ensuring a high human upper bound. In the initial data creation phase, we observed several sources of confusion among human raters, typically related to some negative pairs being frequently labeled as positive; we identified two causes of this discrepancy and then mitigated it through data selection heuristics. 

First, some common monosemous words still might get linked to multiple different Wikipedia pages, thus creating confusing negative pairs. For instance, some pronouns  (e.g., `he', `it') and common nouns (e.g., `daughter', `son') may link to different entities as a result of coreference resolution. However, truly ambiguous words are typically directly defined in Wikipedia Disambiguation pages. We thus keep only the negative pairs that link to separate entries found in the Wikipedia Disambiguation pages. The second issue concerns concept granularity, as Wikipedia sometimes makes too fine-grained distinctions between concepts: e.g., by setting up separate pages for a country's name in different time periods.\footnote{E.g., \textit{`China'} can be linked to the page \textit{`Republic of China (1912–1949)'} and to the page \textit{`Empire of China (1915–1916)'}.} We mitigate this issue by requiring that the negative pairs do not share common or parent Wikipedia categories.

The application of these heuristics during data creation (see \S\ref{s:datacreation}) yields a substantial boost in human performance: e.g., the scores increase from 74\% to 88\% for ZH, and from 76\% to 94\% for DE.

\subsection{Adversarial Examples \label{sec:add challenge}}
Another requirement is assessing to which extent models can grasp the meaning of a target word based on the (complex) \textit{interaction} with its context. However, recently it was shown that SotA pretrained LMs exploit superficial cues while solving language understanding tasks due to spurious correlations seeping into the datasets \citep{gururangan-etal-2018-annotation,niven-kao-2019-probing}. This hinders generalizations beyond the particular datasets and makes the models brittle to minor changes in the input space \citep{jia-liang-2017-adversarial,iyyer-etal-2018-adversarial}. As verified later in \S\ref{s:results}, we found this to be the case also for the existing WiC datasets: just considering the target word and neglecting the context (or vice versa) is sufficient to achieve high performance.

To remedy this issue, we already ensured that models could not rely solely on target words in \S\ref{s:datacreation} by including both positive and negative examples for each ambiguous word in different contexts. Further, we now introduce adversarial negative examples in \resource to penalize models that rely only on context without considering target words. To create such negative examples, we sample a positive pair ${x}_i$ and instead of the original target word $w_i$, we take another related word $\tilde{w}_i$ in the same context $c_i$ as the new target word $w_i$. 

We define the related word as a hyperlinked mention sharing the same parent Wiki category as the original target word: e.g., in Table~\ref{examp} we change the target word `~\begin{CJK*}{UTF8}{gbsn}
阿波罗\end{CJK*}' ({Apollo}) from Example 1 into the related word `\textit{NASA}', resulting in Example 3. Both words share a common Wiki parent category, \begin{CJK*}{UTF8}{gbsn}美国国家航空航天局\end{CJK*} (NASA). 
The contexts of both examples deal with spaceships; hence, only a fine-grained understanding of lexical differences between the target words warrants the ability to recognize \textit{Apollo} as identical to \begin{CJK*}{UTF8}{gbsn}`
阿波罗'\end{CJK*} but different from `\textit{NASA}'. Overall, adversarial examples amount to roughly $1/4$ of our dataset.

\subsection{Data Statistics and Language Coverage}
\begin{table}
\centering
\def\arraystretch{0.93}
{\small
\resizebox{\linewidth}{!}{
\begin{tabular}{lrrrr}
\toprule
&\textbf{WiC} & \textbf{XL-WiC}&\textbf{MCL-WiC}&\textbf{\resource} \\
\cmidrule(lr){2-5}
examples (mean) &7,466&{\bf 14,510}&3600&{13,074} \\
examples (median) &7,466&1,676&2000&{\bf 8,570} \\	
word types (mean) &4,130&7,255& 2766&{\bf 9,868} \\
word types (median) &4,130&1,201& 2072&{\bf 8,520} \\
context length&17&22.7&26.13&{\bf 53.5}\\
\cmidrule(lr){2-5}
human accuracy&80&81.8&-&{\bf 90.6}\\
human agreement&80&- \protect{\footnote{There is only one annotator for most languages in XL-WiC. Therefore no agreement score can be computed.}}&{\bf 94.2}& 88.4 \\
\cmidrule(lr){2-5}
languages&1&12&5&{\bf 15}\\
language families&1&5&3&{\bf 10}\\
\bottomrule

\end{tabular}
}
}%
\vspace{-0.5mm}
\caption{Comparison of the most salient data statistics of \resource versus WiC, XL-WiC and MCL-WiC. 
}
\vspace{-0.5mm}
\label{table:data statistics}
\end{table}
We summarize the main properties of \resource while comparing against previous word-in-context datasets WiC, XL-WiC and MCL-WiC in Table~\ref{table:data statistics}. More detailed per-language scores are listed in Table~\ref{table:allres}. First, we emphasize the accrued reliability of \resource, as both human accuracy and inter-annotator agreement are substantially higher than with WiC and XL-WiC (i.e., rising by \textasciitilde10 points). Second, for a comparable overall dataset size we increase the number of examples and word types in resource-poor languages. If we consider their median across languages, \resource has 8,570 and 8,520, respectively, around four times more than XL-WiC (1,676 and 1,201) and MCL-WiC (2000 and 2072). XL-WiC are heavily skewed towards a small number of languages, namely German and French, and provides large datasets in those languages. MCL-WiC only offers training data for English. In contrast, \resource provides a more balanced representation of its languages. Third, in \resource we deliberately include longer contexts. While the data in WiC, XL-WiC and MCL-WiC are derived from concise dictionary examples, \resource data reflect natural text where key information may be spread across a much wider context.

Our selection of languages is guided by the recent initiatives to cover a typologically diverse language sample \cite{ponti-etal-2020-xcopa}. In particular, \resource covers 15 languages, more than XL-WiC (12 languages) and MCL-WiC (5 languages). Diversity can be measured along multiple axes, such as family, geographic areas, and scripts \citep{ponti2019modeling}. \resource includes 10 language families, namely: Afro-Asiatic (1 language), Austronesian (1), Basque (1), Indo-European (5), Japonic (1), Kartvelian (1), Koreanic (1), Sino-Tibetan (1), Turkic (2), Uralic (1). This provides a more balanced sample of the cross-lingual variation compared to XL-WiC (5 families) and MCL-WiC (3 families). Regarding geography, in addition to the areas covered by XL-WiC and MCL-WiC (mostly Europe and Eastern Asia), we also represent South-East Asia (with ID), the Middle East (TR), the Caucasus (KA), the Indian subcontinent (UR and BN), as well as central Asia (KK). Finally, \resource also introduces scripts that were absent in other datasets, namely the Georgian alphabet and the Bengali script (a Northern Indian abugida), for a total of 8 distinct scripts.

\section{Experimental Setup}
\label{s:exp}
We now establish a series of baselines on \resource to measure the gap between current SotA models and human performance. 


\vspace{1.3mm} \noindent
\textbf{Pretrained Encoders.} Multilingual contextualized representations $\mathbf{e}_i \in \mathbb{R}^d$ for each target word are obtained via \textsc{base} variants of cased multilingual BERT \cite[\textsc{mBERT,}][]{devlin2018bert} and \textsc{XLM-R}\footnote{Due to limited resources, we leave experiments with large variants of the models for future research. } \cite{conneau-etal-2020-unsupervised}, available in the HuggingFace repository \cite{Wolf:2019hf}.


\vspace{1.4mm} \noindent
\textbf{Classification.} Given two contextualized representations $\mathbf{e}_{i,src}$ and $\mathbf{e}_{i,trg}$ for a pair of target words, two setups to make prediction are considered: the first, {\bf metric}-based, is a non-parametric setup. In particular, we follow \citet{pilehvar-camacho-collados-2019-wic} and score the distance $\delta$ between the representations via cosine similarity. A threshold $t$ from the development set is set via grid search across $0.02$ intervals in the $[0, 1]$ interval. Therefore, if $\delta(\mathbf{e}_{i,src}, \mathbf{e}_{i,trg}) \geq t$ the pair is classified as negative, and positive otherwise. On the other hand, the {\bf fine-tuning} setup is parametric: following \citet{raganato-etal-2020-xl}, we train a logistic regression classifier that takes the concatenation of the contextualized representations $[\mathbf{e}_{i,src} \oplus \mathbf{e}_{i,trg}]$ as input.\footnote{The first WordPiece is always used if a word is segmented into multiple WordPieces.} The entire model (both the encoder and the classifier) is then fine-tuned to minimize the cross-entropy loss of the training set examples with Adam \cite{kingma2014adam}. We perform grid search for the learning rate in $[5e-6, 1e-5, 3e-5]$, and train for 20 epochs selecting the checkpoint with the best performance on the dev set. 

\vspace{1.4mm} \noindent
\textbf{Cross-lingual Transfer.} 
In addition to {\bf supervised} learning, we also carry out cross-lingual transfer experiments where data splits may belong to different language pairs. The goal is transferring knowledge from a source \textit{language pair} $\ell_s$ to a target \textit{language pair} $\ell_t$. To simulate different scenarios of data paucity, in the fine-tuning setup we consider: 1) {\bf zero-shot} transfer, where train and development sets belong to $\ell_s$ and the test set to $\ell_t$; 2) {\bf zero-shot + TLD}\footnote{TLD stands for ``Target Language Development (set)''} transfer, which is similar except for the dev set given in $\ell_t$; 3) on top of zero-shot + TLD, we provide a small amount of training examples in $\ell_t$, which we denote as {\bf few-shot} transfer; 4) finally, in the {\bf joint multilingual} setup, we train a single model on the concatenation of the train sets for all language pairs and select the hyper-parameters on the development set of $\ell_t$.

\section{Results and Discussion}
\label{s:results}

\begin{table*}
\centering
\def\arraystretch{0.83}
\small{
\begin{tabular}{clcccccccccccccc}
\toprule
&&DE&	RU&	JA&	ZH&	AR	&KO	&FI&	TR&	ID&	EU&	KA&	BN	&KK&	UR\\
\cmidrule(lr){3-16}
\multirow{2}{*}{\rotatebox[origin=c]{90}{\textsc{Mtr}}} & \textsc{mBERT}&67.1&	65.0 &	62.3&	65.8&	63.9&	62.1&	61.7&	57.1&	\textbf{66.3}&	\textbf{64.1}&	\textbf{60.4}&	\textbf{60.0}&	\textbf{59.2}&	\textbf{58.8}\\
& \textsc{XLM-R}&65.0 	&63.1 &	56.7&	56.7&	58.4&	57.5&	64.1&	62.4&	65.7&	62.9&	58.3&	56.2&	58.0&	55.5\\
\cmidrule(lr){3-16}
\multirow{2}{*}{\rotatebox[origin=c]{90}{\textsc{Ft}}} & \textsc{mBERT}& \textbf{80.0} &	77.4 &	73.9&	\textbf{71.0}&	\textbf{67.4}&	\textbf{68.2}&	\textbf{71.6}&	\textbf{69.3}&	64.6&	62.2 & --&--&--&--\\
& \textsc{XLM-R}&77.4 &\textbf{76.1} &	\textbf{75.9}&	68.9&	65.9&	65.3&	68.4&	64.4&	54.6&	55.8 &--&--&--&--\\
\cmidrule(lr){3-16}
\cmidrule(lr){3-16}
\multirow{2}{*}{\rotatebox[origin=c]{90}{\textsc{Hm}}} & accuracy &93.5&	89.5&93.0&	87.5&	93.5&	93.5&	90.5&	90.5&	91.5&92.5&	90.0&89.5&85.5&88.0\\
& agreement &90.0&78.0&90.0&94.0	&100.0&	92.0	&88.0&	96.0	&92.0&	84.0&	94.0&	80.0&	80.0&	80.0	\\
\bottomrule
\end{tabular}}
\vspace{-1mm}
\caption{Accuracy of \textsc{mBERT} and \textsc{XLM-R} on \resource in a supervised learning setting. We report metric-based classification (\textsc{Mtr}) results, as well as the scores in the fine-tuning setup (\textsc{Ft}). The third group of rows (\textsc{Hm}) displays human performance, in terms of both accuracy and inter-rater agreement. Results for the larger test sets for DE and RU are reported in Table~\ref{table:allres larger} in the Appendix.}
 \label{table:allres}
 \vspace{-1.5mm}
\end{table*}

\noindent \textbf{Metric-based vs Fine-tuning.}
We report the results for the supervised learning setting (where all data splits belong to the same language pair) in Table~\ref{table:allres}. The metric-based approach achieves consistent scores across all languages, fluctuating within the range $[57.1, 67.1]$ for \textsc{mBERT} and $[55.5, 65.0]$ for \textsc{XLM-R}. This indicates that the pretrained encoder alone already contains some relevant linguistic knowledge, to a certain degree. In comparison, fine-tuning yields more unequal results, being more data-hungry. In particular, it performs worse than the metric-based approach on languages with small training data size (e.g., ID and EU in Table~\ref{table:allres}), whereas it surpasses the metric-based approach on languages with abundant examples (e.g., DE, RU). 

\vspace{1.4mm}
\noindent \textbf{\textsc{XLM-R} vs \textsc{mBERT}.} 
Table~\ref{table:allres} also reveals that \textsc{XLM-R} is more sensitive to train data size than \textsc{mBERT}, often falling behind in both the metric-based and fine-tuning setups, especially for resource-poorer languages. These findings are in line with what \citet{vulic2020multi} report for Multi-SimLex, which are grounded on lexical semantics similarly to \resource. However, they contradict the received wisdom from experiments in other multilingual sentence-level tasks \citep{ponti-etal-2020-xcopa, conneau-etal-2020-unsupervised}, where \textsc{XLM-R} outperforms \textsc{mBERT} in cross-lingual transfer. While the exact causes go beyond the scope of this work, we speculate that the two encoders excel in separate aspects of semantics, the lexical and the sentence level.

\begin{figure}[t]%
\centering{
\small{
\subfigure[\textsc{mBERT}]{
\includegraphics[width=0.4\textwidth]{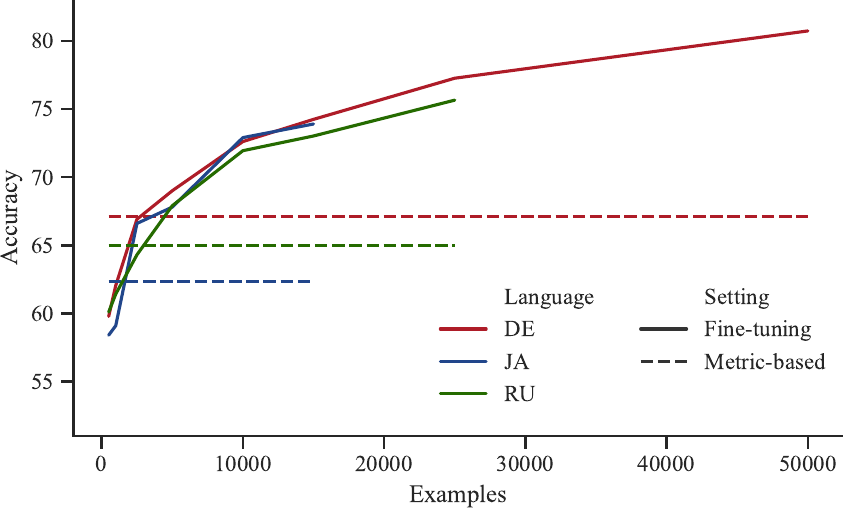}}
\subfigure[\textsc{XLM-R}]{

\includegraphics[width=0.4\textwidth]{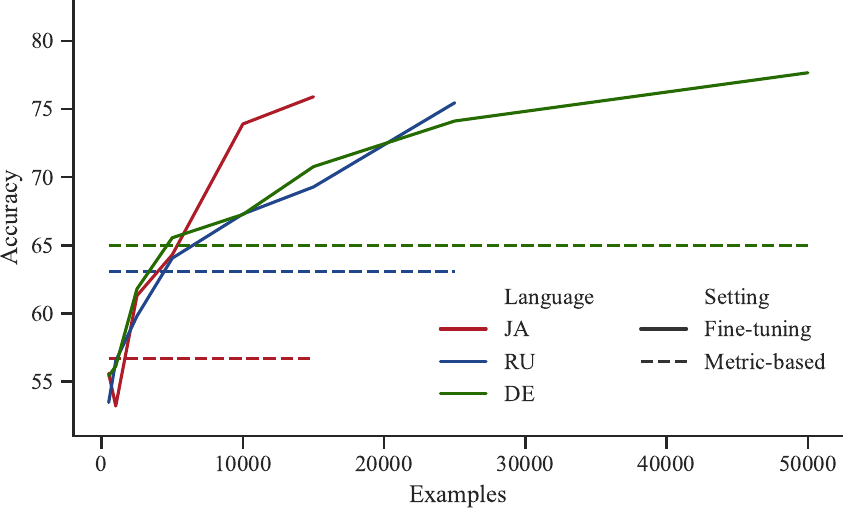}}}
\\

    
    \caption{Impact of train data size (for fine-tuning) on performance in DE, RU (larger test sets used), and JA from \resource. X axis is in the log scale.  \label{fig:de}}}
\end{figure}

\vspace{1.4mm}
\noindent \textbf{Effect of Data Size on Fine-Tuning.} 
\label{app:datasize}
To further investigate the effect of train data size on fine-tuning, we perform an in-depth analysis on some selected languages (DE, RU and JA). Note that we use the larger dev and test sets for DE and RU for this experiment. We study how performance changes as we vary the number of training examples from 500 to the full set.
The results in Figure~\ref{fig:de} indicate that, while fine-tuning starts lower than the metric-based baseline, it grows steadily and begins to take the lead from around 2,500 train examples. 

\vspace{1.3mm}
\noindent \textbf{Zero-shot Transfer.} 
The results are presented in Table~\ref{table:language transfer average}. We select the training data of each of the five languages with most data (DE, RU, JA, ZH, AR) in turn for source-language fine-tuning. Subsequently, we report the average prediction performance across all remaining 9 target languages. 

\begin{table}
\centering
\def\arraystretch{0.85}
\footnotesize{
\begin{tabularx}{\linewidth}{r p{0.07\textwidth} p{0.07\textwidth}p{0.07\textwidth}p{0.07\textwidth}}
\toprule
{} &\multicolumn{2}{c}{\it Zero-shot}&\multicolumn{2}{c}{\it +TLD}\\

$\ell_s$&\textsc{mBERT}&\textsc{XLM-R}&\textsc{mBERT}&\textsc{XLM-R}\\
\cmidrule(lr){2-5}
DE (all)  &71.2&{\bf 72.0}&71.5&71.7\\
RU (all) &{71.1}&69.8&71.0&69.9\\
JA (all) &68.1&61.9&{68.6}&63.2\\
ZH (all) &66.2&60.3&{66.6}&62.1\\
AR (all) &{67.7}&61.8&67.1&62.1\\
\cmidrule(lr){2-5}
DE (10k)&65.4&	62.4&	 65.9&	62.5\\
RU (10k)& 66.4&	64.7&	66.1&	64.6\\
JA (10k)& 67.5&	61.3&	67.2	&61.7\\
ZH (10k)& 66.0&	62.8&	65.8&	62.1\\
AR (10k)& {\bf 67.7}&61.8&	67.1&	62.1\\
\bottomrule

\end{tabularx}}
\vspace{-1mm}
\caption{Zero-shot transfer from 5 high-resource source languages to the remaining 9 languages in \resource. The parentheses contain (approximate) train data sizes.}
\label{table:language transfer average}
\vspace{-1.5mm}
\end{table}


First, we note that the TLD variant for hyperparameter selection does not yield gains. Second, the best choice of a source language appears to be German across the board, achieving an average score of 71.2 with \textsc{mBERT} and 72.0 with \textsc{XLM-R}. Nevertheless, this is simply due to its ample number of examples (50k). In fact, when controlling for this variable by equalizing the total size of each train split to 10k, see the bottom half of Table~\ref{table:language transfer average}, all source languages perform comparably. 

Breaking down the average results into individual languages in Table~\ref{table:language transfer} (top section), however, reveals an even more intricate picture. In particular, the best source language for KK is RU and for JA is ZH, rather than DE. This can be explained by the fact that these pairs share their scripts, Cyrillic and Kanji / Hanzi, respectively, at least in part. 
This indicates that a resource-leaner but related language might sometimes be a more effective option as source language than a resource-rich one. It is also noteworthy that zero-shot transfer from DE outperforms supervised learning in most languages, except for those both resource-rich and distant (JA, ZH and AR).

\begin{table*}
\centering
\def\arraystretch{0.7}
\footnotesize{
\begin{tabular}{lcccccccccccccccc}
\toprule
$\ell_s$ &model&DE&	RU&	JA&	ZH&	AR	&KO	&FI&	TR&	ID&	EU&	KA&	BN	&KK&	UR\\
\cmidrule{3-16}
\rowcolor{Gray}
\multicolumn{16}{c}{\bf Zero-shot Transfer (+TLD)}\\
\cmidrule{3-16}
DE&\textsc{mBERT}& -&	{\bf 75.1}&	71.1&	{\bf 68.9}&	64.8&	71.1&	{\bf 78}&	{\bf 75.7}&	{\bf 75.4}&	{\bf 74.2}&	{\bf 70.2}&	68&	62.8 &	{\bf 68.5}\\
&\textsc{XLM-R}&-&	{\bf 77.5}&	70.1&	67.7&	67.4&	70.2&	{\bf 77.2}&	{\bf 76.6}&	{\bf 74.8}&	{\bf 70.7}&	{\bf 71.9}&	{\bf 70.6}&	66.5&	{\bf 67}\\
\cmidrule{3-16}
RU&\textsc{mBERT}&{\bf 74.4}&	-&	{\bf 71.7}&	68.4	&{\bf 65.1}&	71&	74.8&	72.6&	74.5&	71.2&	69.3&	{\bf 70.3}&	{\bf 67}&	67.8&\\
&\textsc{XLM-R}&{\bf 72.4}&	-&	68.2&	64.4&	{\bf 69.5}&	{\bf 72.3}&73.7&	71.2&	69.7&	68.8&	69.6&	68.3&	{\bf 69}&	66.7\\
\cmidrule{3-16}
JA&\textsc{mBERT}&70.9&	70.7&	-&	{\bf 68.9}&	63.2&	{\bf 72.5}&	70.9&	72.5&	70.1&	67.7&	65.3&	67.9&	65.8&	65\\
&\textsc{XLM-R}&64.7&	65.7&	-&	{\bf 70.4}&	63.2&	67.1&	66.5&	64.9&	62.5&	62.5&	61.7&	66.7&	57.8&	59.3\\
\cmidrule{3-16}
ZH&\textsc{mBERT}&71.5&	69.2&	68.7&-&	61.6&	70.6&	68.6&	70.8&	66.9&	68&	64.9&	65.6&	61.5&	62.8\\
&\textsc{XLM-R}&66.5&	67.1&	{\bf 72.1}&	-&	62.6&	62.9&	66.2&	63.8&	63.4&	61.6&	59.8&	61.1&	59.8&	60.3\\
\cmidrule{3-16}
AR&\textsc{mBERT}&69.2&	67.6&	68.5&	66.3&	-&	67.1&	68.6&	69.5&	68.2	&67.6&	66.9&	68.1&	63.8&	63.8\\
&\textsc{XLM-R}&62&	63&	62.2&	62.0&	-&	61.9&	62.2&	63.6	&63.9&	61.3&	62.1&	64.9&	60.8&	58.3\\
\cmidrule{3-16}
\rowcolor{Gray}
\multicolumn{16}{c}{\bf Joint Multilingual Learning}\\
\cmidrule{3-16}
all&\textsc{mBERT}&{\bf 80.4}&	{\bf 82.1}&	78.2&	75.2&	73.3&	75.8&	{\bf 81.2}&	{\bf 80.6}	&{\bf 78.4}&	{\bf 75.9}&	76.5&	{\bf 76}	&71.3&	{\bf 73.3}\\
&\textsc{XLM-R}& 79.4 &		80.9 &	{\bf 79.4}&	{\bf 76.1}&	{\bf 73.6}&	{\bf 76}	&{\bf 81.2}&	80.5&	77.9&	74.2&	{\bf 77.7}&	73&	{\bf 74.5}&	72.5\\
\bottomrule
\end{tabular}}
\vspace{-1.5mm}
\caption{Results for zero-shot transfer from a single source (top section) and joint transfer from multiple sources (bottom section) in \resource. The best scores for each setup are in bold. Results for the larger testsets of DE an RU are reported in Table~\ref{table:language transfer larger}.}
\label{table:language transfer}
\vspace{-1.5mm}
\end{table*}

\begin{figure}[t!]
\centering
\includegraphics[width=0.88\columnwidth]{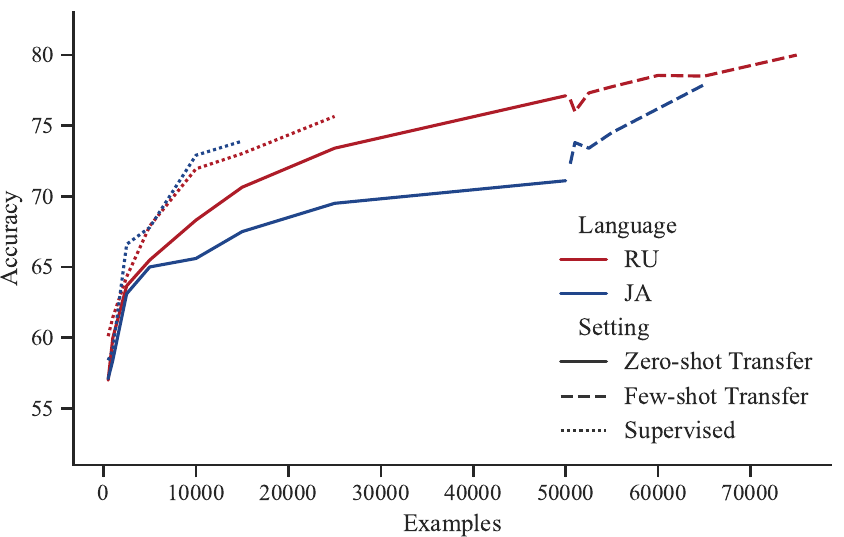}
\vspace{-1mm}
\caption{\textsc{mBERT} performance in two target languages, RU (larger test set) and JA, across different amounts of training examples under different settings: supervised learning (examples directly from $L_{trg}$) few-shot transfer (examples from DE $L_{src}$ plus $L_{trg}$ examples), and zero-shot transfer (+TLD) from DE.
 \label{fig:ru ja}}
    \vspace{-2mm}
\end{figure}

\vspace{1.4mm}
\noindent \textbf{Few-shot Transfer.} 
To study the differences between training on $\ell_s$ and $\ell_t$ with controlled train data size, we plot the model performance on two target languages (RU and JA) as a function of the amount of available examples across different transfer conditions in Figure~\ref{fig:ru ja}. Comparing supervised learning (based on target language data) with zero-shot learning (based on DE data), it emerges how the former is always superior if the number of examples is the same. However, zero-shot learning may eventually surpass the peak performance of supervised learning by taking advantage of a larger pool of examples: this is the case in RU, but not in JA. This illustrates a trade-off between quality (in-domain but possibly scarce data) and quantity (abundant but possibly out-of-domain data).

Few-shot learning combines the desirable properties of both approaches. After pre-training a model on DE, it can be adapted on a small amount of target-language examples. Performance continues to grow with more shots; with as few as 1k JA examples it is comparable to supervised learning on 15k examples. Few-shot learning thus not only achieves the highest scores, but also leverages costly target-language data in a sample-efficient fashion.

\begin{table}[t]
\centering
\def\arraystretch{0.75}
\footnotesize{
\begin{tabularx}{\columnwidth}{l lXXXX}
\toprule
Dataset & $\ell$ & Full & Ctx & Tgt & Hm \\
\midrule
\textbf{WiC} & EN & 67.1 & 65.1 & 55.2 & 80.0 \\
\textbf{XL-WiC} & DE & 81.0 & 75.0 & 80.0 & 74.0 \\
\textbf{MCL-WiC}& EN & 83.3 & 82.6 & 53.4 & 96.8\\
\cmidrule(lr){2-6}
\textbf{\resource -A}& DE & 84.2 & 83.6 & 50.0 & 93.0 \\
\textbf{\resource +A}& DE & 80.0 & 73.6  & 66.9 & 93.5 \\
\textbf{\resource -A}& ZH & 79.8 & 78.2 & 49.0 & 86.0 \\
\textbf{\resource +A}& ZH & 71.0 & 66.0 & 61.6 & 87.5 \\
\bottomrule

\end{tabularx}}
\vspace{-1.5mm}
\caption{The impact of adversarial examples on \textsc{mBERT} performance. +A indicates the presence of adversarial examples, -A their absence. \textsc{Ctx} represents training a model on the context only, \textsc{Tgt} on the target word only, and \textsc{full} on the whole input. \textsc{Hm} stands for human accuracy.}
\vspace{-2mm}
\label{table:context vs target word}
\end{table}

\vspace{1.4mm}
\noindent \textbf{Joint Multilingual Learning.}
The results are shown in Table~\ref{table:language transfer}. We observe a substantial boost in performance across all the languages compared to both zero-shot transfer from any individual language and supervised learning (cf.\ Table~\ref{table:allres}), including high-resource languages such as DE and RU. Low-resource languages enjoy the most copious gains: with \textsc{mBERT} as the encoder, UR improves by 4.8 points, KK by 4.3, BN by 5.7, and KA by 6.3. However, this is still insufficient to equalize performances across the board, as the latter group of languages continues to lag behind: between DE and UR remains a gap of 7.2 points. We speculate that the reason behind this asymmetry is the fact that in addition to being resource-poor, UR, KK, BN, and KA are also typologically distant from languages where most of the examples are concentrated. Overall, these findings suggest that leveraging multiple sources is better than a single one by virtue of the transfer capabilities of massively multilingual encoders, as previously demonstrated \citep{wu2019beto,ponti2020parameter}.



\vspace{1.4mm}
\noindent \textbf{Adversarial Examples.}
\label{ssec:advres}
Finally, we investigate whether the inclusion of adversarial examples (see \S\ref{sec:add challenge}) makes \resource less likely to be solved by relying on superficial clues. In Table~\ref{table:context vs target word}, we compare the performance of \textsc{mBERT} trained on the full input (which we label \textsc{Full}) with two adversarial baselines. 
We implement the \textsc{Tgt} variant by inputting only the target word to the classification model and the \textsc{Ctx} variant by replacing the target word with a `\textit{[MASK]}' token. We perform analysis across \resource, WiC, XL-WiC and MCL-WiC.\footnote{We select languages with sufficient and comparable training data: ZH for \resource, DE for \resource and XL-WiC and EN for MCL-WiC.}

In previous datasets, at least one of the adversarial baselines reaches performance close to the \textsc{full} model: in WiC (EN), \textsc{Ctx} has a gap of only 2 points. In XL-WiC (DE), \textsc{Tgt} is only 1 point away from \textsc{Full}. In MCL-WiC (EN), the gap between \textsc{Ctx} and \textsc{Full} is even below 1 point. This would also be the case in \resource were it not for the extra adversarial examples (rows +A): by virtue of this change, the distance between \textsc{Full} and the best adversarial baseline is 6.4 points in DE and 5.0 in ZH. Therefore, it is safe to conclude that a higher score on \resource better reflects a deep semantic understanding by the model. Moreover, the last column of Table~\ref{table:context vs target word} also includes reference human accuracy. While the best baseline in XL-WiC even surpasses the human upper bound, the addition of adversarial examples in \resource combined with higher human accuracy drastically increases the gap between the two: 13.5 points in DE and 16.5 in ZH. Overall, this results in a much more challenging evaluation benchmark.

In addition, we report separate results on these adversarial examples, and compare with model performance on non-adversarial examples for DE, RU, JA, ZH and AR, within the supervised setting (Table~\ref{table:adver}). It is clear that the adversarial examples pose a much greater challenge for the model with overall much lower scores. This is expected as for adversarial examples the models must have a finer-grained and more accurate understanding of both the target word semantics and the surrounding (sentential) context. 


\begin{table}[t]
\centering
\def\arraystretch{0.75}
\footnotesize{
\begin{tabularx}{\columnwidth}{l lXXXX}
\toprule
 & DE & RU & JA & ZH & AR\\
\midrule
Adversarial &73.7	&66.5 &	65.3 &	64.7&	55.0 \\
Non-adversarial &83.1&77.9&	76.5&	75.5&70.5 \\
\bottomrule

\end{tabularx}}
\vspace{-1.5mm}
\caption{\textsc{mBERT} performance on adversarial and non-adversarial examples across five datasets in \resource.}
\vspace{-2mm}
\label{table:adver}
\end{table}

\section{Related Work}
\textbf{Cross-Lingual Evaluation of Word Meaning in Context.}
Going beyond readily available sense inventories required for WSD-style evaluations, the comprehensive benchmarks for evaluating word meaning in context cross-lingually are still few and far between. XL-WiC \cite{raganato-etal-2020-xl} extends the original English WiC framework of \citet{pilehvar-camacho-collados-2019-wic} to 12 other languages, but supports only monolingual evaluation, and suffers from issues such as small gaps between human and system performance. The SemEval-2021 shared task MCL-WiC does focus on cross-lingual WiC, but covers only five high-resource languages from three language families (English, French, Chinese, Arabic, Russian). Both XL-WiC and MCL-WiC mainly focus on common words and do not include less frequent concepts (e.g., named entities). Further, their language coverage and data availability are heavily skewed towards Indo-European languages.


There are several other `non-WiC' datasets designed to evaluate cross-lingual context-aware lexical representations. Bilingual Contextual Word Similarity (BCWS) \cite{bcws} challenges a model to predict graded similarity of cross-lingual word pairs given sentential context, one in each language. In the Bilingual Token-level Sense Retrieval (BTSR) task \cite{liu-etal-2019-investigating}, given a query word in a source language context, a system must retrieve a meaning-equivalent target language word within a target language context.\footnote{BTSR could be seen as a contextualized version of the standard bilingual lexicon induction task \cite[\textit{inter alia}]{Mikolov:2013arxiv,Sogaard:2018acl,Ruder:2019jair}.} However, both BCWS and BTSR are again very restricted in terms of language coverage: BCWS covers only one language pair (EN-ZH), while BTSR contains two pairs (EN-ZH/ES). Further, they provide only test data: as such, they can merely be used as general intrinsic probes for pretrained models, but cannot support fine-tuning experiments and cannot fully expose the relevance of information available in pretrained models for downstream applications. This is problematic as intrinsic tasks in general do not necessarily correlate well with downstream performance \cite{chiu-etal-2016-intrinsic, glavas-etal-2019-properly}. 


\vspace{1.3mm}
\noindent \textbf{\resource vs. Entity Linking.}
Our work is related to the entity linking (EL) task \cite{rao2013entity, cornolti2013framework,shen2014entity} similarly to how the original WiC (based on WordNet knowledge) is related to WSD. EL systems must map entities in context to a predefined knowledge base (KB). While WSD relies on the WordNet sense inventory, the EL task focuses on KBs such as Wikipedia and DBPedia. When each entity mention is mapped to a unique Wiki page, this procedure is termed wikification \cite{mihalcea2007wikify}. The cross-lingual wikification task \cite{ji2015overview, tsai-roth-2016-cross} grounds multilingual mentions to English Wikipedia pages. Similar to WSD, EL evaluation is tied to a specific KB. It thus faces similar limitations of WSD in terms of restricting meanings and their distinctions to those predefined in the inventory. In comparison, \resource leverages Wikipedia only as a convenient resource for extracting the examples, similar to how the original WiC work leverages WordNet. \resource itself is then framed on natural text, without requiring the modeling of the KBs. Also, in comparison with EL, \resource provides higher data quality and a more challenging evaluation of complex word-context interactions, achieved by a carefully designed data extraction and filtering procedure.  


\section{Conclusion}
We presented \resource, a large-scale and challenging multilingual benchmark for evaluating word meaning in context (WiC) across languages. \resource is constructed by leveraging multilingual Wikipedias, and subsequently validated by humans. It covers 15 typologically diverse languages and a vocabulary substantially larger than all previous WiC datasets. As such, it provides more comprehensive and reliable quality estimates for multilingual encoders. Moreover, \resource includes adversarial examples: resolving such examples requires genuine lexical understanding, as opposed to relying on spurious correlations from partial input. Finally, \resource offers the possibility to perform cross-lingual evaluation, pairing context between different languages. Moreover, we explored the impact of language relatedness on model performance by transferring knowledge from multiple source languages. We established a series of baselines on \resource, based on SotA multilingual models, revealing that the task is far from being `solved' even with abundant training data. All models struggle especially when transferring to distant and resource-lean target languages. 
We hope that \resource will guide and foster further research on effective representation learning across different languages.

\section*{Acknowledgments}

We thank the anonymous reviewers
for their helpful feedback. We acknowledge Peterhouse College at University of Cambridge for funding Qianchu Liu's PhD research. The work was also supported by the ERC Consolidator Grant LEXICAL: Lexical Acquisition Across Languages (no 648909) awarded to Anna Korhonen. We also appreciate many helpful discussions and feedback from our colleagues in the Language Technology Lab.
\bibliography{anthology,custom}
\bibliographystyle{acl_natbib}

\clearpage
\appendix

\section{Results on the larger test set in DE and RU}

Table~\ref{table:allres larger} and Table~\ref{table:language transfer larger} list results for the larger test sets of DE and RU in \resource.

\begin{table}[h]
\centering
\def\arraystretch{0.83}
\small{
\begin{tabular}{clcccccccccccccc}
\toprule
&&DE large&	RU large&\\
\cmidrule(lr){3-4}
\multirow{2}{*}{\rotatebox[origin=c]{90}{\textsc{Mtr}}} & \textsc{mBERT}&66.1&	65.3 \\
& \textsc{XLM-R}&64.5 	&63.2\\
\cmidrule(lr){3-4}
\multirow{2}{*}{\rotatebox[origin=c]{90}{\textsc{Ft}}} & \textsc{mBERT}& 80.7	&75.7 \\
& \textsc{XLM-R}&77.7&	75.5\\

\bottomrule
\end{tabular}}
\caption{Accuracy of \textsc{mBERT} and \textsc{XLM-R} on larger test sets of DE and RU in \resource in a supervised learning setting. We report metric-based classification (\textsc{Mtr}) results, as well as the scores in the fine-tuning setup (\textsc{Ft}).}
 \label{table:allres larger}
\end{table}

\begin{table}[h]
\centering
\def\arraystretch{1.0}
\footnotesize{
\begin{tabular}{lcccc}
\toprule
$\ell_s$ &model&DE large&	RU large\\
\cmidrule{3-4}
\rowcolor{Gray}
\multicolumn{4}{c}{\bf Zero-shot Transfer (+TLD)}\\
\cmidrule{3-4}
DE&\textsc{mBERT}& -&	77.1\\
&\textsc{XLM-R}&-&	76.4\\
\cmidrule{3-4}
RU&\textsc{mBERT}& 76.1&	-\\
&\textsc{XLM-R}&72.5&	-\\
\cmidrule{3-4}
JA&\textsc{mBERT}&72.3&	72.2\\
&\textsc{XLM-R}&65.5&	66.0\\
\cmidrule{3-4}
ZH&\textsc{mBERT}&70.8&	70.0\\
&\textsc{XLM-R}&64.7&	64.4\\

\cmidrule{3-4}
AR&\textsc{mBERT}&69.1&	68.9\\
&\textsc{XLM-R}&62.9 &	63.7\\
\cmidrule{3-4}
\rowcolor{Gray}
\multicolumn{4}{c}{\bf Joint Multilingual Learning}\\
\cmidrule{3-4}
all&\textsc{mBERT}& 81.9& 81.2\\
&\textsc{XLM-R}& 80.5&	80.7 \\
\bottomrule
\end{tabular}}
\caption{\resource transfer results for the larger test sets of DE and RU. Performance from zero-shot transfer from a single source is in the top section; joint transfer from multiple sources is in the bottom section.}
\label{table:language transfer larger}
\end{table}

\end{document}